\documentclass{article}

\usepackage{microtype}
\usepackage{graphicx}
\usepackage{amsmath}
\usepackage{subcaption}
\usepackage{booktabs} 

\usepackage{hyperref}

\usepackage[preprint]{icml2026}

\usepackage{amssymb}
\usepackage{mathtools}
\usepackage{amsthm}
\usepackage{algorithmic}
\usepackage{booktabs} 
\usepackage{enumitem}
\usepackage{multirow}
\usepackage{array}
\usepackage[table]{xcolor}
\newcommand{\methodname}{KPO}
\usepackage[table]{xcolor}
\definecolor{cRed}{HTML}{D62728}
\definecolor{cGreen}{HTML}{2CA02C}
\definecolor{cBlue}{HTML}{1F77B4}
\definecolor{cOrange}{HTML}{FF7F0E}

\usepackage[capitalize,noabbrev]{cleveref}

\theoremstyle{plain}

\theoremstyle{definition}

\theoremstyle{remark}

\usepackage{pifont}

\usepackage[textsize=tiny]{todonotes}

\icmltitlerunning{Online Causal Kalman Filtering for Stable and Effective Policy Optimization}

\begin{document}

\twocolumn[
  \icmltitle{Online Causal Kalman Filtering for Stable and Effective Policy Optimization}




  \begin{icmlauthorlist}
    \icmlauthor{Shuo He}{yyy}
    \icmlauthor{Lang Feng}{yyy}
    \icmlauthor{Xin Cheng}{yyy}
    \icmlauthor{Lei Feng}{comp}
    \icmlauthor{Bo An}{yyy}
  \end{icmlauthorlist}

  \icmlaffiliation{yyy}{Nanyang Technological University, Singapore. Email: \url{shuohe123@gmail.com}}
  \icmlaffiliation{comp}{Southeast University, China}

  \icmlcorrespondingauthor{Lei Feng}{fenglei@seu.edu.cn}


  \vskip 0.3in
]



\printAffiliationsAndNotice{}  

\begin{abstract}
Reinforcement learning for large language models suffers from high-variance token-level importance sampling (IS) ratios, which would destabilize policy optimization at scale. To improve stability, recent methods typically use a fixed sequence-level IS ratio for all tokens in a sequence or adjust each token's IS ratio separately, thereby neglecting temporal off-policy derivation across tokens in a sequence. In this paper, we first empirically identify that \emph{local off-policy deviation is structurally inconsistent at the token level}, which may distort policy-gradient updates across adjacent tokens and lead to training collapse. To address the issue, we propose Online Causal \underline{K}alman Filtering for stable and effective \underline{P}olicy \underline{O}ptimization (\methodname{}). Concretely, we model the desired IS ratio as a latent state that evolves across tokens and apply a Kalman filter to update this state online and autoregressively based on the states of past tokens, regardless of future tokens. The resulting filtered IS ratios preserve token-wise local structure-aware variation while strongly smoothing noise spikes, yielding more stable and effective policy updates. Experimentally, \methodname{} achieves superior results on challenging math reasoning datasets compared with state-of-the-art counterparts. Code is available at \url{https://github.com/shuohe1995/verl-kpo}
\end{abstract}

\section{Introduction}

Reinforcement learning (RL) has become a key method for advancing large language models (LLMs) beyond the limits of pretraining~\citep{openai2024learning,yang2025qwen3,deepseekai2025deepseek_r1}. With large-scale RL, e.g., GRPO~\citep{deepseekai2025deepseek_r1}, LLMs can acquire substantially stronger long-horizon reasoning ability, enabling solutions to challenging tasks across diverse domains, e.g., code generation~\citep{lyu2025let}, information retrieval~\citep{zhang2025agentorchestra}, software engineering~\citep{qian2024chatdev}, and open-ended device control~\citep{tan2024cradle,bai2025digi}.

\begin{figure}[!t]
    \centering
    \includegraphics[width=1.0\linewidth]{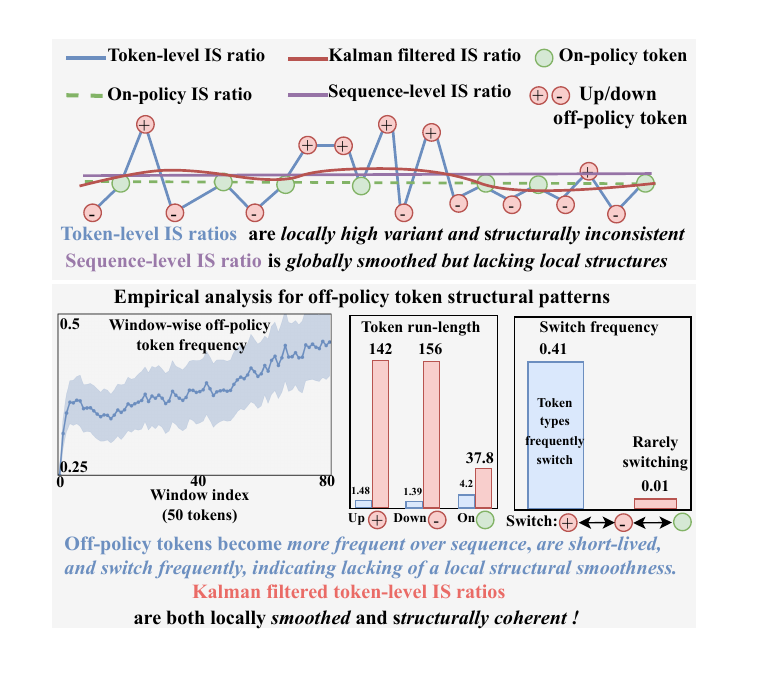}
    \vspace{-0.1in}
    \caption{Illustration of local structural off-policy patterns. Raw token-level importance-sampling (IS) ratios (blue) exhibit high local variance and structural inconsistency, whereas a sequence-level IS ratio (purple) is globally smooth but obscures within-sequence structure. Off-policy frequency increases over the sequence (window-wise statistics), off-policy runs are short-lived (run-length distribution), and token states switch frequently, suggesting weak local coherence. Token-level Kalman filtering (red) yields locally smoothed yet structurally consistent IS ratios.}
    \label{fig:observation}
\end{figure}

Despite these gains, recent research reveals that GRPO-style optimization can become unstable at scale, with high policy-gradient variance and entropy collapse~\citep{cui2025entropy}. Moreover, instability is often worsened in off-policy settings (e.g., mini-batch updates), where the policy gradient additionally depends on an importance-sampling (IS) ratio between the updated policy and the behavior (old) policy, potentially amplifying policy-gradient variance~\citep{zheng2025stabilizing}. In practice, the IS ratio can be further destabilized by Mixture-of-Experts routing discontinuities~\citep{zheng2025group}, train-inference mismatches~\citep{liu2025speed}, and inconsistent numerical precision~\citep{qi2025defeating}. Consequently, reducing the variance of the IS ratio is critical for stabilizing large-scale policy optimization~\citep{zheng2025stabilizing}. To this end, recent methods typically replace token-wise ratios with a fixed \emph{sequence-level} IS ratio, e.g., GSPO~\citep{zheng2025group} and GMPO~\citep{zhao2025geometric}, or separately adjust token-wise IS ratios via soft gating functions~\citep{gao2025soft} or flipping the IS ratios of positive-advantage tokens~\citep{wang2025aspo}. However, they totally neglect the temporal structure of off-policy derivation across tokens within a sequence, which is important for stable and effective policy optimization. 

Intuitively, off-policy deviation across tokens within a sequence should exhibit \emph{global heterogeneity} but \emph{local homogeneity}. For instance, in complex mathematical reasoning, a single response typically comprises multiple distinct reasoning paths, which may differ substantially in off-policy deviation, i.e., global heterogeneity. Yet within a specific reasoning path, adjacent tokens share similar local semantics and should therefore display coherent, slowly varying off-policy deviation, i.e., local homogeneity.

Motivated by this intuition, in this paper, we pioneer a new temporal perspective on token-wise IS ratios and examine the patterns of off-policy derivation. Specifically, we perform GRPO on Qwen3-4B and record corresponding token-wise IS ratios. The deatiled experimental setting is shown in Appendix~\ref {app:off-policy}. We design three complementary statistics for off-policy tokens: (i) window-wise occurrence frequency, (ii) run-lengths (consecutive occurrences), and (iii) switching frequency (categorized into \texttt{up} when ratio $>1$ and \texttt{down} when ratio $<1$). As shown in Figure~\ref{fig:observation}, window-wise occurrence frequency increases along the sequence, suggesting that more off-policy tokens occurs in the later part of the sequence. Meanwhile, off-policy tokens are short-lived (low run-lengths), and states switch frequently (e.g., switch frequency $\approx 0.41$), indicating weak local structural coherence. Overall, these observations point to a key phenomenon: \emph{local off-policy deviation is structurally inconsistent at the token level.} The phenomenon counters our earlier intuition of local homogeneity and could distort the updates of policy gradients across adjacent tokens and ultimately results in the training collapse of GRPO.

To address this issue, we propose \emph{Online Causal \underline{K}alman Filtering for stable and effective \underline{P}olicy \underline{O}ptimization} (\methodname{}), a structure-aware autoregressive smoothing method for token-wise IS ratios. Our key idea is to model the token-wise IS ratio as a time series within a state-space framework: a process model captures smooth latent dynamics across adjacent tokens, while an observation model accounts for noise in the computed ratios. We then apply an online Kalman filter to produce a token-wise filtered ratio using only the information from past and current tokens. The estimated IS ratio of the current token will consider the local past tokens' IS ratios, thereby preserving the local off-policy structure. In this way, the Kalman-filtered IS ratio is both numerically smoothed and structurally coherent. As shown in Figure~\ref{fig:observation}, the Kalman-filtered IS ratio (red) increases the token run-lengths and decreases the token switching, which enables more stable and more effective policy optimization. Extensive experiments on six challenging mathematical reasoning tasks validate the stability and effectiveness of \methodname{} compared with current state-of-the-art methods. 

Our main contributions are:
\begin{itemize}[leftmargin=0.4cm,topsep=-1pt]
    \item \textbf{\emph{An empirical revealing for local off-policy patterns.}} We empirically reveal the structural inconsistency of local off-policy derivation in token-level IS ratios, which could degrade the policy optimization.
    \item \textbf{\emph{A novel policy optimization method: \methodname{}.}} We propose causal Kalman filtering over token ratios to smoothing local fluctuations while preserving within-sequence local coherent structure, improving stability and effectiveness under off-policy updates.
    \item \textbf{\emph{Strong and stable empirical results.}} \methodname{} consistently improves performance on math-reasoning benchmarks over state-of-the-art baselines, while simultaneously enhancing training stability.
\end{itemize}

\section{Related Work}

\noindent\textbf{Post-training for LLMs.}\quad
Supervised fine-tuning~\citep{wei2022finetuned,sanh2022multitask} adapts pretrained LLMs to instruction following via demonstrations. In contrast, RL from human feedback (RLHF)~\citep{stiennon2020learning,ouyang2022training,bai2022training,Kpo2024,garg2025ipo} optimizes a preference reward model with KL regularization to a reference policy. TRPO~\citep{schulman2015trust} imposes an explicit KL constraint, while PPO~\citep{schulman2017proximal} uses a clipped surrogate and remains the default for scalable RLHF. Recent post-training reduces critic and on-policy overhead. DPO~\citep{rafailov2023direct} yields a closed-form objective for KL-regularized preference learning, replacing RL with supervised optimization. GRPO~\citep{shao2024deepseekmath,shao2025deepseekmath} further removes the critic by using group-relative normalization over multiple samples per prompt, reducing memory while retaining PPO-style~\citep{zheng2023secrets,touvron2023llama} stability. Beyond single-turn alignment, RL has also been extended to long-horizon agentic settings~\citep{feng2025group,zhou2023language,zeng2023agenttuning,zhai2024fine,yao2023react,zhang2024codeagent,chen2025reinforcement} 

\noindent\textbf{Advanced variants of group-based RL.}\quad
Recent group-based RL methods extend GRPO-style objectives by refining how learning signals are estimated, normalized, and utilized. 
A first line of work focuses on improving advantage estimation and group weighting to correct biased or degenerate updates. 
Dr.\ GRPO~\citep{liu2025understanding} identifies estimator-driven collapse in \textsc{R1-zero}-like training, while GVPO~\citep{zhang2025gvpo} derives principled group weights from KL-constrained reward maximization to achieve better variance control. 
Building on more reliable advantage estimates~\citep{yang2025not}, subsequent work addresses instability introduced by normalization under bounded or multi-reward supervision. 
BNPO~\citep{xiao2025bnpo} applies Beta-based normalization to stabilize gradients on bounded rewards, whereas GDPO~\citep{liu2026gdpo} shows that shared normalization can collapse advantages in multi-reward settings and resolves this issue through reward-decoupled normalization. 
As group-based methods are increasingly applied to long-chain-of-thought rollouts, several studies target training stability and efficiency at scale. 
DAPO~\citep{yu2025dapo} and Stable-GRPO~\citep{dai2025stable} introduce practical stabilizers such as decoupled clipping~\citep{yang2025dcpo} and staged truncation to mitigate length-induced collapse, while CPPO~\citep{lin2025cppo} and ARM~\citep{wu2025arm} improve sample efficiency by pruning low-contribution completions or adapting reasoning formats on a per-instance basis. 
When external rewards become unreliable or unavailable, recent work explores uncertainty-aware updates. 
SEED~\citep{chen2025seed} and RIGHT~\citep{zhang2025right} leverage model uncertainty for reweighting or intrinsic supervision, whereas minimalist analyses~\citep{xiong2025minimalist} and KRPO~\citep{wang2025kalman} clarify the role of filtering effects and smoothing in group-based advantage estimation.

\noindent\textbf{IS-ratio-oriented RL.}\quad
To achieve stable policy optimization at scale, GMPO~\citep{zhao2025geometric} and GSPO~\citep{zheng2025group} use fixed sequence-level ratios for all tokens within a sequence. In contrast, SAPO~\citep{gao2025soft} introduces adaptive clipping via soft gating functions to continuously modulate token-wise ratios based on their magnitudes. Besides, ASPO~\citep{wang2025aspo} employs asymmetric ratio handling to tighten the update on risky high-ratio tokens while retaining learning signal elsewhere. 

\section{Preliminaries}
\label{sec:preliminaries}

In this section, we introduce the necessary notions for policy optimization. We consider a dataset $\mathcal{D}$ of prompt-response pairs $(x,y)$, where the response $y=[y_1,y_2,\ldots,y_T]$ with $T$ tokens is generated autoregressively by a large language model parameterized by $\theta$, i.e., a policy $\pi_\theta:\pi_\theta(y\mid x)=\prod_{t=1}^T \pi_\theta(y_t\mid x,y_{<t})$. A prompt-response pair $(x,y)$ can be assigned a score $s(x,y)\in[0,1]$ by a verifier $\mathcal{R}$.

\noindent\textbf{Proximal policy optimization (PPO).}\quad
We optimize a new policy $\pi_\theta$ using samples collected from an old policy $\pi_{\theta_{\text{old}}}$. For each token $y_t$, we define the token-level importance-sampling (IS) ratio
$r_t=\frac{\pi_{\theta}(y_t\mid x,y_{<t})}{\pi_{\theta_{\text{old}}}(y_t\mid x,y_{<t})}$, which corrects for the distribution shift between the behavior policy and the updated policy. PPO uses a token-level advantage estimate $A_t$, typically obtained from a learned value function (e.g., via GAE~\citep{schulman2015high}). The clipped surrogate objective of PPO is
\begin{equation}
\label{eq:J-ppo-grpo-ppo}
\mathcal{J}_{\text{PPO}}=
\mathbb{E}\!\left[
\frac{1}{T}\sum_{t=1}^{T}
\min\!\Big(
r_t\,A_t,\,
r^{\prime}_t\,A_t
\Big)
\right],
\end{equation}
where $r^{\prime}_t=\operatorname{clip}\!\big(r_t,\,1-\epsilon,\,1+\epsilon\big)$ and $\epsilon$ controls the clipping range. The clipping operation limits overly large policy updates by capping the effective ratio used in the objective.

\noindent\textbf{Group relative policy optimization (GRPO).}\quad
GRPO follows the same clipped form, but avoids an explicit value model. Instead, for each prompt $x$, it samples a group of $G$ responses $\{y_i\}_{i=1}^{G}$ from $\pi_{\theta_{\text{old}}}$, evaluates each response with a verifier score $s(\cdot,\cdot)$, and constructs a sequence-level relative advantage. This advantage is then shared across all tokens within the corresponding response: $\widehat{A}_{i,t}=
\frac{s(x_i,y_i)-\operatorname{mean}\!\big(\{s(x_i,y_i)\}_{i=1}^{G}\big)}
{\operatorname{std}\!\big(\{s(x_i,y_i)\}_{i=1}^{G}\big)}.$ The resulting objective of GRPO is
\begin{equation}
\small
\label{eq:J-ppo-grpo-grpo}
\mathcal{J}_{\mathrm{GRPO}}=
\mathbb{E}\!\left[
\frac{1}{G}\sum_{i=1}^{G}
\frac{1}{T}\sum_{t=1}^{T}
\min\!\Big(
r_{i,t}\,\widehat A_{i,t},\,
r^{\prime}_{i,t}\,\widehat A_{i,t}
\Big)
\right],
\end{equation}

\noindent\textbf{Token-level policy gradient in GRPO.}\quad
To better understand how the IS ratio affects the policy gradient, we formulate the policy gradient of the GRPO objective for an individual token:
\begin{equation}
    r_{i,t}\,\nabla_\theta \log \pi_\theta\!\left(y_{i,t}\mid x, y_{i,<t}\right)\widehat A_{i,t}.
\end{equation}
In the on-policy setting, the IS ratio $r_{i,t}$ is equal to 1 for all tokens, while in the off-policy setting, the IS ratio could vary among the clipping bounds, which could have a negligible impact on the policy gradient, such as amplifying the covariance effect of tokens with high probability and high advantages~\citep{cui2025entropy}. Hence, GRPO could be unstable for policy optimization at scale~\citep{zheng2025stabilizing}, especially under the off-policy optimization.   

\noindent\textbf{Sequence-level IS ratios.}\quad
To address the issue, GSPO~\citep{zheng2025group} and GMPO ~\citep{zhao2025geometric} directly use a fixed sequence-level IS ratio (stopped gradient) for each token in a sequence, by averaging token-wise IS ratios across the entire sequence:
\begin{equation}
\overline r_i
=\left(\prod_{t=1}^{T} r_{i,t}\right)^{\frac{1}{T}}
=\exp\!\left(\frac{1}{T}\sum_{t=1}^{T}\log r_{i,t}\right).
\label{eq:seq_ratio_product}
\end{equation}
This practice reduces variance by totally smoothing token-wise fluctuations into a single scalar, regardless of the within-sequence off-policy heterogeneity. In contrast, we focus on local structure-aware smoothing for token-wise IS ratios.

\section{Method: Online Causal Kalman Filtering}
\begin{figure}[!t]
    \centering
    \includegraphics[width=1.0\linewidth]{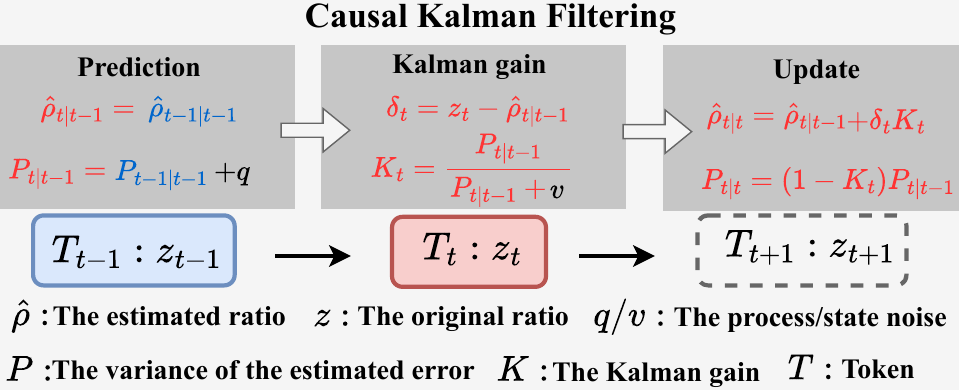}
    \vspace{-0.1in}
    \caption{Causal Kalman filtering. The filter alternates prediction, adaptive gain computation, and update to produce a smoothed estimate $\hat{\rho}_{t|t}$ and its uncertainty $P_{t|t}$ from streaming observations $z_t$, using process noise $Q$ and observation noise $V$.}
    \label{fig:kpo}
\end{figure}

In this section, we will first clarify our motivation and then provide a detailed introduction to our proposed method, as shown in Figure~\ref{fig:kpo}, which mainly includes three processes to autoregressively filter the IS ratio.

\subsection{Motivation}
As discussed above, using sequence-level importance sampling (IS) ratios is an effective method for stabilizing policy optimization at scale. However, this stability assumes that all tokens in a sequence have the same degree of off-policy derivation, which is commonly violated in policy optimization. For instance, in mathematical reasoning tasks, the response generally includes different reasoning steps in a chain-of-thought, along which the updated policy may only drift from the behavior policy on certain reasoning steps but maintain on-policy on others. 

In contrast, our intuition is that although off-policy degrees can vary across \emph{semantic segments}, they should exhibit \emph{local coherence} within a segment because neighboring tokens are usually semantically consistent. It is therefore implausible for the policy to alternate sharply or frequently relative to the behavior policy at the token-to-token level.

These considerations motivate us to design a method that can not only smooth the IS ratio but also preserve the local coherent structure. For this purpose, we treat per-token IS ratios as a structured but noisy time series over token positions and seek an online estimate of the latent, locally coherent off-policy signal using only past and current tokens, corresponding to the autoregressive generation of tokens.

\subsection{Kalman filtering for token-wise IS ratios}
Formally, we treat $\{z_{i,t}\}_{t=1}^{T}$ as a left-to-right observed time series for the sequence $x_i$ including $T$ tokens. In particular, it is associated with an underlying \emph{latent} smoothed IS ratio $\rho_{i,t}$. Notably, we consider the causal Kalman filtering in the log space of the IS ratios $z_{i,t}=\text{log}r_{i,t}$ due to numerical stability. Our objective is to estimate the latent IS ratio $\rho_{i,t}$ by Kalman filtering, which is both locally smoothed and structurally coherent. Next, we will first introduce the Kalman state-space model, and then detail the filtering process, including three processes.

\noindent\textbf{Kalman state-space model.}\quad
To characterize the variation of the latent (observed) IS ratio, given each sequence $x$ including $T$ tokens (we omit the subscript $i$ for simplicity in the next), we model the latent and observed IS ratio~\citep{grewal2025kalman} respectively:
\begin{equation}
\left\{
\begin{aligned}
\rho_{t} &= \rho_{t-1} + \eta_{t}, \quad \eta_{t}\sim \mathcal{N}(0,Q),\\
z_{t} &= \rho_{t} + \epsilon_{t}, \quad \epsilon_{t}\sim \mathcal{N}(0,V).
\end{aligned}
\right.
\label{eq:state_obs_model}
\end{equation}
where $\rho_{t}$ is the latent desired IS ratio and $Q/V$ is the process/observation noise variance.$Q$ and $V$ govern the Kalman filter’s smoothing tracking trade-off. $Q$ is the \emph{process-noise} variance, controlling how quickly the latent IS ratio may evolve across tokens. A larger $Q$ permits faster drift and improves responsiveness to real local shifts, while a smaller $Q$ imposes stronger temporal smoothness at the cost of potential lag. In contrast, $V$ is the \emph{observation-noise} variance, reflecting the reliability of the observed IS ratio $z_{t}$. A larger $V$ down-weights noisy observations and suppresses isolated spikes and rapid switching, whereas a smaller $V$ makes the estimate follow $z_{t}$ more closely. Together, $Q$ and $V$ determine the Kalman gain and thus balance denoising against fidelity to true off-policy dynamics in \methodname{}.

\noindent\textbf{\scalebox{1.2}{\ding{172}} Prediction step.}\quad
To capture the intuition that the off-policy degree may evolve gradually across token positions, we employ a naive random-walk dynamics with process-noise variance $Q$ controlling how quickly the latent signal is allowed to drift. Formally, let $\hat\rho_{t\mid t}$ and $P_{t\mid t}$ denote the posterior mean and variance of $\rho_{t}$ given observations $z_{1:t}$. Similarly, $\hat\rho_{t\mid t-1}$ and $P_{t\mid t-1}$ denote the one-step-ahead predictive quantities given $z_{1:t-1}$. We initialize with a prior $(\hat\rho_{0\mid 0},P_{0\mid 0})$ (e.g., $\hat\rho_{0\mid 0}=0$ and $P_{0\mid 0}=P_0$). $\hat\rho_{0\mid 0}=0$ indicates the IS ratio of the first token is on-policy (1). The prediction step will only use the observed IS ratio of past tokens, regardless of future tokens, corresponding to the autoregressive generation of tokens:
\begin{equation}
\left\{
\begin{aligned}
\hat\rho_{t\mid t-1} &= \hat\rho_{t-1\mid t-1},\\
P_{t\mid t-1} &= P_{t-1\mid t-1} + Q.
\end{aligned}
\right.
\label{eq:kalman_pred}
\end{equation}

\noindent\textbf{\scalebox{1.2}{\ding{173}} Kalman gain.}\quad
After predicting the posterior mean and variance of $\rho_{t}$, the next step is to predict the Kalman gain, which is the key for Kalman filtering. The Kalman gain determines how the filter trades off the current observation against the prior prediction when estimating the latent IS ratio at token position $t$. Given the one-step prediction $(\hat\rho_{t\mid t-1}, P_{t\mid t-1})$, we first compute the innovation (residual)
\begin{equation}
\delta_{t} = z_{t} - \hat\rho_{t\mid t-1},
\end{equation}
which measures the discrepancy between the observed log-ratio and the predicted latent value. The Kalman gain is then
\begin{equation}
K_{t} = \frac{P_{t\mid t-1}}{P_{t\mid t-1}+V},
\label{eq:kalman_gain}
\end{equation}
By construction, $K_{t}\in[0,1]$ in the scalar case and acts as an \emph{adaptive step size} for correcting the prediction with the new measurement. When the predicted uncertainty $P_{t\mid t-1}$ is large (or the measurement noise $V$ is small), $K_{t}$ increases and the filter places more weight on $z_{t}$, allowing rapid adaptation to real changes in off-policy degree. Conversely, when $V$ is large, $K_{t}$ decreases and the update relies more on the temporal prior, thereby suppressing isolated spikes and high-frequency fluctuations in $z_{t}$. In our setting, this adaptive weighting is crucial, since it preserves persistent, structure-aware shifts in token-wise off-policy dynamics while attenuating system-induced noise in the observed ratios.

\noindent\textbf{\scalebox{1.2}{\ding{174}} Update step.}\quad
Given the innovation $\delta_{t}$ and the Kalman gain $K_{t}$, we incorporate the new observation by performing a linear correction of the one-step prediction:
\begin{equation}
\left\{
\begin{aligned}
\hat\rho_{t\mid t} &= \hat\rho_{t\mid t-1}+K_{t}\,\delta_{t},\\
P_{t\mid t} &= (1-K_{t})\,P_{t\mid t-1}.
\end{aligned}
\right.
\label{eq:update_block}
\end{equation}
The first line performs an innovation-based correction: $K_{t}$ serves as an adaptive step size that controls how far the estimate moves from the prior $\hat\rho_{t\mid t-1}$ toward the measurement $z_{t}$. Larger $K_{t}$ makes the filter track genuine changes more quickly, while smaller $K_{t}$ favors the temporally smoothed prior and suppresses transient fluctuations. The second line updates uncertainty: incorporating $z_{t}$ reduces the posterior variance by a factor $(1-K_{t})$, with higher gains implying greater information gain. In our setting, this recursion stabilizes token-wise weighting by damping isolated spikes (large $\delta_{t}$ with small $K_{t}$) while allowing persistent deviations across neighboring tokens to accumulate into smooth, coherent shifts in $\hat\rho_{t\mid t}$.

\subsection{The policy optimization objective of \methodname{}}
After filtering in log space, we map the latent estimate back to ratio space by exponentiation:
$\widetilde r_{t}=\exp(\hat\rho_{t\mid t})$. We then replace the raw token-wise IS ratio $r_{t}$ in GRPO-style objectives with the filtered ratio $\widetilde r_{t}$. Let $\widehat A_{t}$ denote a token-shaped sequence-level advantage. The resulting \methodname{} objective is
\begin{equation}
\small
\label{eq:J_kpo}
\mathcal{J}_{\mathrm{\methodname{}}}
=
\left\{
\begin{aligned}
&\mathbb{E}\!\left[
\frac{1}{G}\sum_{i=1}^{G}
\frac{1}{T}\sum_{t=1}^{T}
\min\!\Big(
\widetilde r_{i,t}\,\widehat A_{i,t},\,
\widetilde r^{\prime}_{i,t}\,\widehat A_{i,t}
\Big)
\right],\\
&\mathbb{E}\!\left[
\frac{1}{G}\sum_{i=1}^{G}
\frac{1}{T}\sum_{t=1}^{T}
\widetilde r_{i,t}\,\widehat A_{i,t}
\right],
\end{aligned}
\right.
\end{equation}
where $\widetilde r^{\prime}_{t}=\operatorname{clip}\!\big(\widetilde r_{t},\,1-\epsilon^{-},\,1+\epsilon^{+}\big)$ and $\epsilon^{-}$($\epsilon^{+}$) is the clipping threshold. In particular, the first line corresponds to the clipped surrogate, while the second line recovers the unclipped variant. Compared with sequence-level aggregation (Eq.~\eqref{eq:seq_ratio_product}), \methodname{} preserves token-wise heterogeneity in off-policy degree by maintaining per-token ratios, yet substantially reduces the variance through causal filtering. As a result, \methodname{} inherits the fine-grained credit assignment of token-level reweighting while improving optimization stability. The full procedure is summarized in Algorithm~\ref{alg:kpo} in Appendix~\ref{app:algorithm}.

\begin{table*}[!t]
\centering
\caption{\textbf{Main results on challenging math reasoning tasks.} Columns are benchmarks and rows are methods. We report \texttt{avg@16} and \texttt{pass@16}. Higher is better. Best results are in \textbf{bold} and second-best are \underline{underlined}. \methodname{ achieves overall superior performance.}}
\label{tab:main_result_math_reformat}
\setlength{\tabcolsep}{3.5pt}
\renewcommand{\arraystretch}{1.10}
\resizebox{\textwidth}{!}{%
\begin{tabular}{l
|cc 
|cc 
|cc 
|cc 
|cc 
|cc 
}
\toprule
\multirow{2}{*}{\textbf{Method}} &
\multicolumn{2}{c|}{\textbf{AIME'24}} &
\multicolumn{2}{c|}{\textbf{AIME'25}} &
\multicolumn{2}{c|}{\textbf{AMC'23}} &
\multicolumn{2}{c|}{\textbf{MATH500}} &
\multicolumn{2}{c|}{\textbf{Minerva}} &
\multicolumn{2}{c}{\textbf{Olympiad}} \\
& avg@16 & pass@16
& avg@16 & pass@16
& avg@16 & pass@16
& avg@16 & pass@16
& avg@16 & pass@16
& avg@16 & pass@16\\
\midrule
GRPO
& 27.29 & 53.33 & 23.12 & 43.33 & 73.43 & 92.50 & 85.66 & 92.80 & \underline{38.67} & 50.36 & 48.60 & 62.25 \\
GMPO
& 30.83 & 50.00 & 27.50 & 46.66 & 76.56 & 87.50 & 86.62 & 93.40 & 38.48 & \textbf{50.73} & 49.27 & 60.92 \\
GSPO
& 32.70 & 60.00 & 29.16 & \underline{50.00} & 75.46 & \underline{95.00} & 87.41 & \underline{94.00} & 37.17 & 47.79 & 51.37 & \underline{63.00}  \\ \midrule

\cellcolor{gray!15}\methodname{}-unclipped
& \cellcolor{gray!15}\textbf{34.79} & \cellcolor{gray!15}\textbf{66.67} & \cellcolor{gray!15}\textbf{33.75} & \cellcolor{gray!15}\textbf{50.00} & \cellcolor{gray!15}\textbf{80.00} & \cellcolor{gray!15}\textbf{95.00} & \cellcolor{gray!15}\textbf{88.24} & \cellcolor{gray!15}93.00  & \cellcolor{gray!15}\textbf{39.15}  & \cellcolor{gray!15}\underline{50.36} & \cellcolor{gray!15}\textbf{52.31} & \cellcolor{gray!15}62.55 \\

\cellcolor{gray!15}\methodname{}-clipped
& \cellcolor{gray!15}\textbf{37.91} & \cellcolor{gray!15}\textbf{63.33} & \cellcolor{gray!15}\textbf{36.87} & \cellcolor{gray!15}\textbf{60.00} & \cellcolor{gray!15}\textbf{87.50} & \cellcolor{gray!15}\textbf{95.00} & \cellcolor{gray!15}\textbf{89.42} & \cellcolor{gray!15}\textbf{94.80} & \cellcolor{gray!15}38.23 & \cellcolor{gray!15}48.16 & \cellcolor{gray!15}\textbf{54.06} & \cellcolor{gray!15}\textbf{66.27}  \\

\bottomrule
\end{tabular}%
} 
\end{table*}

\begin{figure*}[!t]
    \centering
    \includegraphics[width=1.0\linewidth]{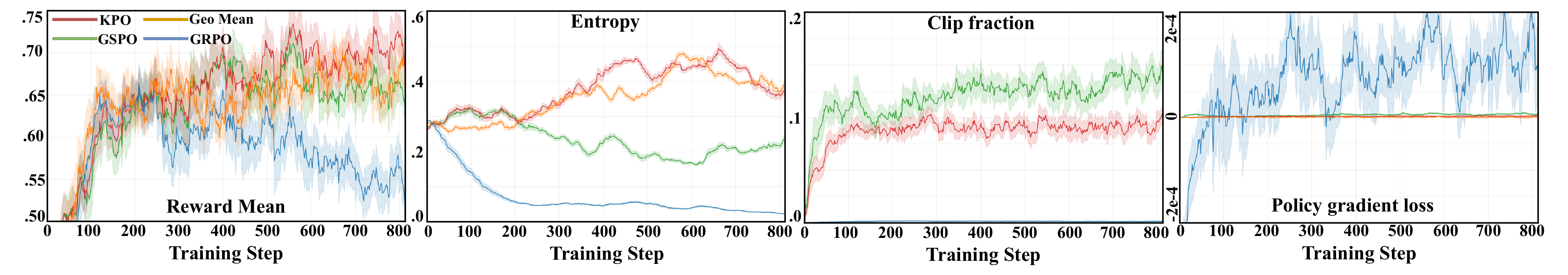}
       \vspace{-0.1in}
    \caption{\textbf{Training dynamics of \methodname{}-clipped over optimization steps}. From left to right: mean episodic reward, policy entropy, PPO clip fraction, and policy gradient loss. Solid lines denote the average across runs, and the shaded bands indicate variability across runs.}
    \label{fig:training}
\end{figure*}

\section{Experiments}
In this section, we describe the evaluation setting, baselines, and implementation details for assessing \methodname{} on challenging mathematical reasoning benchmarks.

\subsection{Setup}
We use Qwen3-4B~\citep{yang2025qwen3} for evaluation. Unless stated otherwise, all methods share the same model initialization, tokenizer, and inference pipeline. For RL fine-tuning, we adopt the supervised RL training corpus from DAPO~\citep{yu2025dapo}, which contains diverse mathematical problems paired with verifiable solutions and corresponding reward signals.

\noindent\textbf{Evaluation benchmarks and metrics.}\quad
We follow the standard protocol for math-reasoning evaluation and report \texttt{avg@16} and \texttt{pass@16} on five competitive benchmarks: AIME'24, AIME'25, AMC'23, MATH500~\citep{hendrycks2021measuring}, and OlympiadBench. For each problem, we sample $16$ candidate solutions and compute: (i) \texttt{pass@16}, the fraction of problems for which at least one sample is correct, and (ii) \texttt{avg@16}, the mean accuracy across the 16 samples.
The benchmarks are: \textbf{AIME'24}, AIME 2024 problems that emphasize contest-style multi-step reasoning with short numeric answers;
\textbf{AIME'25}, AIME 2025 problems with the same answer format and comparable difficulty;
\textbf{AMC'23}, AMC 2023 multiple-choice problems targeting intermediate contest mathematics;
\textbf{MATH500}, a 500-problem subset of MATH covering broad topics with multi-step reasoning and solution supervision; and
\textbf{OlympiadBench}, Olympiad-level problems designed to stress high-difficulty reasoning and proof-oriented skills.

\noindent\textbf{Baselines.}\quad
We compare \methodname{} with token-level GRPO and representative sequence-level ratio stabilization methods, including GMPO and GSPO. All methods are evaluated under the same inference protocol, model backbone, and decoding budget to ensure a controlled comparison.

\noindent\textbf{Implementation details.}\quad
For a fair comparison, we adopt the same training configuration for all methods: a training batch size of 32, a minibatch size of 8 (yielding four off-policy minibatches), and a group size of 8. we set the Kalman filter parameters to $Q=1\times10^{-6} (V=1)$ for \methodname{}-clipped and $Q=1\times10^{-4} (V=1)$ for \methodname{}-unclipped respectively. The clipping parameters $\epsilon^{-}$ and $\epsilon^{+}$ are set to 0.0003 and 0.0004 for \methodname{}-clipped. All methods use the same sampling strategy and generate $16$ candidate solutions per problem for computing \texttt{avg@16} and \texttt{pass@16}. The detailed setting is shown in Appendix~\ref{app:exp_set}.

\begin{table*}[t!]
\centering
\caption{Comparison of token-level dynamics before and after Kalman filtering (KF). Reported are the token proportion of \emph{Up/Down/On}, the mean run-length for each token state, the overall token switching frequency, the low-frequency ratio (LFR), and the global and windowed local variance. All reported values are averaged over all samples.}
\setlength{\tabcolsep}{3.5pt}
\renewcommand{\arraystretch}{1.0}
\resizebox{1.0\textwidth}{!}{%
\begin{tabular}{c|ccc|ccc|c|c|cc}
\toprule
\multirow{2}{*}{\textbf{Metrics}} 
& \multicolumn{3}{c|}{\textbf{Token proportion}} 
& \multicolumn{3}{c|}{\textbf{Token run-length}} 
& \multirow{2}{*}{\textbf{Token Switch frequency}} 
& \multirow{2}{*}{\textbf{Low frequency ratio}} 
& \multicolumn{2}{c}{\textbf{Variance}} \\

& \textbf{Up} & \textbf{Down} & \textbf{On} 
& \textbf{Up } & \textbf{Down } & \textbf{On } 
& &

& \textbf{Glob. var.} & \textbf{Win. loc. var.} \\
\midrule
\textbf{Before KF}     & 0.25 & 0.22 & 0.53 & 1.64 & 1.57 & 3.53  & 0.43 (high) & 0.12 (low) & 0.19 & 0.15 \\
\rowcolor{gray!15}\textbf{After KF}   & 0.35 & 0.43 & 0.22 & 119.95 & 135.12 & 35.11 & 0.01 (low) & 0.98 (high) & 1e-4 & 1e-5 \\
\bottomrule
\end{tabular}
}
\label{tab:raw_vs_filtered_metrics_grouped_transposed}
\end{table*}
\subsection{Experimental results}
\noindent\textbf{\methodname{} achieves overall superior performance.}\quad
Table~\ref{tab:main_result_math_reformat} summarizes \texttt{avg@16} and \texttt{pass@16} on six math benchmarks. Overall, \methodname{} consistently improves over strong baselines: GRPO, GMPO, and GSPO, demonstrating the effectiveness of Kalman filtering. In particular, \methodname{}-clipped achieves the best \texttt{avg@16} on five benchmarks: AIME'24 (37.91), AIME'25 (36.87), AMC'23 (87.50), MATH500 (89.42), and Olympiad (54.06), and also attains the best \texttt{pass@16} on AIME'25 (60.00), MATH500 (94.80), and Olympiad (66.27). On AMC'23, \methodname{} matches the top \texttt{pass@16} (95.00, tied with GSPO) while substantially improving \texttt{avg@16}. Meanwhile, \methodname{}-unclipped attains the best \texttt{pass@16} on AIME'24 (66.67) and the best \texttt{avg@16} on Minerva (39.15). Minerva is the only benchmark where \methodname{} does not lead in \texttt{pass@16}. GMPO is slightly higher (50.73 vs.\ 50.36), while \methodname{} remains competitive and improves \texttt{avg@16}. Overall, these experimental results indicate that \methodname{} achieves consistent gains across benchmarks, which definitely validates its superiority.

\noindent\textbf{\methodname{} improves most on the challenging benchmarks.}\quad
The gains are most pronounced on AIME'24 and AIME'25, where long-horizon reasoning and compounding errors make stable optimization especially critical. Compared with the strongest baseline GSPO, \textbf{\methodname{}-clipped} improves AIME'24 from 32.70/60.00 to 37.91/63.33 in \texttt{avg@16}/\texttt{pass@16}, and \textbf{\methodname{}-unclipped} further raises \texttt{pass@16} to 66.67. On AIME'25, \textbf{\methodname{}-clipped} yields a larger margin, improving GSPO from 29.16/50.00 to 36.87/60.00. These results suggest that smoothing token-wise IS ratios while preserving local structure is particularly beneficial for challenging multi-step reasoning.

\noindent\textbf{\methodname{}-clipped is generally stronger than \methodname{}-unclipped.}\quad
Across benchmarks, the clipped variant consistently outperforms the unclipped one, with the largest gains on AIME'24, AIME'25, and AMC'23 (Minerva being the main exception). A potential explanation is that Kalman filtering smooths ratio noise but does not fully remove truly extreme, structurally off-policy tokens. PPO-style clipping can still suppress the influence of these tokens, effectively preventing locally coherent off-policy segments from dominating the update. This behavior contrasts with standard GRPO, where clipped events tend to be sporadic and locally incoherent, and with GSPO (GMPO), where clipping is driven by a global sequence-level ratio. These results suggest that combining \methodname{} with clipping yields a better bias-stability trade-off than filtering alone. 

\noindent\textbf{Training dynamics.}\quad
We also show the training dynamics of frou methods over optimization steps in Figure \ref{fig:training}. The metrics are Reward means, Entropy, Clip fraction, and Policy gradient loss. The specific meanings of these metrics are shown in Appendix~\ref{app:train_metric}. From Figure~\ref{fig:training}, we can summarize the following observations:  

\begin{itemize}[leftmargin=0.4cm,topsep=-1pt]
    \item \textbf{Reward mean.} All curves rise rapidly in the first $\sim$100 steps. Afterwards, \textcolor{cRed}{\methodname{}} continues to improve and finishes with the highest reward. \textcolor{cGreen}{GSPO} and \textcolor{cOrange}{GMPO} peak mid-training and then plateau. \textcolor{cBlue}{GRPO} diverges after $\sim$200 steps and steadily degrades, ending far below the others. Overall, \textcolor{cBlue}{GRPO} is clearly unstable. \textcolor{cGreen}{GSPO} (\textcolor{cOrange}{GMPO}) are stable but saturate early. \textcolor{cRed}{\methodname{}} is both stable and effective throughout training.

    \item \textbf{Entropy.} \textcolor{cBlue}{GRPO} collapses to near-zero entropy early, indicating near-deterministic behavior and curtailed exploration. In contrast, \textcolor{cRed}{\methodname{}} and \textcolor{cOrange}{GMPO} maintain relatively high entropy with mild fluctuations, while \textcolor{cGreen}{GSPO} stabilizes at a lower but non-zero level. This suggests \textcolor{cRed}{\methodname{}} mitigates entropy collapse and better preserves exploration.

    \item \textbf{Clip fraction.} The clip fractions of \textcolor{cRed}{\methodname{}} and \textcolor{cGreen}{GSPO} increases sharply early and then stabilize. In contrast, \textcolor{cOrange}{GMPO} and \textcolor{cBlue}{GRPO} always exhibit near-zero clip fractions (around $0.0015$). This is because their clipping bounds are set to be larger than the former. Notably, \textcolor{cRed}{\methodname{}} clips fewer tokens than \textcolor{cGreen}{GSPO}. This is because the Kalman-filtered IS ratios are locally structurally coherent, and only part of the tokens within a sequence could be clipped, while the sequence-level IS ratios could result in the overall clipping towards the whole sequence.    

    \item \textbf{Policy gradient loss.} \textcolor{cBlue}{GRPO} shows large, noisy oscillations with high variance, consistent with unstable optimization. The other methods remain small policy loss with low variability, indicating more controlled and stable updates.
\end{itemize}

\subsection{Distribution of Kalman-filtered IS ratios}
In this section, we will further analyze the distribution of the Kalman-filtered IS ratio as shown in Table~\ref{tab:raw_vs_filtered_metrics_grouped_transposed} in terms of (i) the fraction of tokens assigned to each state (Up, Down, On), (ii) how long each state persists once it appears (run-length), (iii) how often the state changes between adjacent tokens (switch frequency), and (iv) frequency- and variance-based measures that reflect whether the sequence is dominated by fast fluctuations or slow trends. The detailed calculation of these metrics is shown in Appendix~\ref{app:exp_set}.

\noindent\textbf{Token proportion.}\quad
This metric measures how often each token state occurs in the sequence. Before filtering, the distribution is dominated by the On state (0.53), with Up and Down appearing at similar rates (Up: 0.25, Down: 0.22). After Kalman filtering, the mass shifts away from On (0.22) and toward the two directional states (Up: 0.35 and Down: 0.43). This shift could not be viewed as undesirable for policy optimization. Although the proportion of on-policy tokens decreases, their local off-policy deviation is structurally smooth and coherent as shown in the token run-length and switching-frequency statistics. More detailed analysis is shown in Appendix~\ref{app:token_pro}.

\noindent\textbf{Token run-length.}\quad
Run-length is the average number of consecutive tokens that stay in the same state. It captures temporal persistence: short run-lengths imply fragmented, noisy state assignments, while long run-lengths imply stable segments. Before Kalman filtering, all states had short runs (Up: 1.64, Down: 1.57, On: 3.53), suggesting frequent fragmentation and many short-lived excursions. After Kalman filtering, run-lengths increase dramatically (Up: 119.95, Down: 135.12, On: 35.11). The filtered sequence, therefore, forms long contiguous blocks, especially for Up and Down, which is consistent with Kalman filtering enforcing temporal coherence across neighboring tokens.

\noindent\textbf{Token switch frequency.}\quad
Switch frequency measures how often the state changes between adjacent tokens (normalized by sequence length). It is a direct measure of discontinuity or mixing among states: higher values mean frequent alternation, lower values mean fewer boundaries and more clustering. Before Kalman filtering, the switch frequency is high (0.43), indicating that the state changes roughly every few tokens. After Kalman filtering, it drops sharply to 0.01, meaning that boundaries between states become rare and state assignments are strongly clustered. This supports the observation from run-length: Kalman filtering greatly reduces short-range alternation and merges nearby tokens into consistent segments.

\noindent\textbf{Frequency-domain view.}\quad
Compared to the view of temporal structure, we follow the view of the frequency domain to analyze the IS ratio. We define \emph{the low-frequency ratio} to measure how much of the signal energy lies in low-frequency components. Low LFR implies rapid, high-frequency fluctuations, while high LFR implies slow variations over long spans of tokens. Before Kalman filtering, LFR is very small (0.12), consistent with a signal dominated by high-frequency changes. After KF, LFR rises to 0.98, showing that the filtered ratio sequence is almost entirely explained by low-frequency components. This aligns with the role of Kalman filtering as a temporal filter that suppresses fast oscillations and retains slow trends. The details of frequency-domain anlysis is shown in Appendix~\ref{app:frequency}.

\noindent\textbf{Global and local variance.}\quad
Variance quantifies the magnitude of fluctuations in the ratio signal. The global variance measures variability over the full sequence, while the windowed local variance measures variability within short windows, capturing local instability. We can see that before Kalman filtering, both are non-trivial (global: 0.19; local: 0.15), indicating substantial fluctuations both overall and within short spans. After Kalman filtering, both variances drop to near zero ($\mathrm{1e-4}$ and $\mathrm{1e-5}$), implying that the filtered IS ratio is nearly constant at both global and local scales. Together with the higher LFR and the reduced switch frequency, this confirms that Kalman filtering strongly dampens local volatility and produces a much more stable token-level IS ratios.

Overall, \methodname{} transforms the token dynamics from a rapidly switching, high-frequency pattern into long, coherent segments with minimal local variation. This distribution shift of the IS ratios will straighten the policy gradients of local tokens within a sequence and thus achieves the stable and effective policy optimization.

\begin{figure}[!t]
    \centering
    \includegraphics[width=1.0\linewidth]{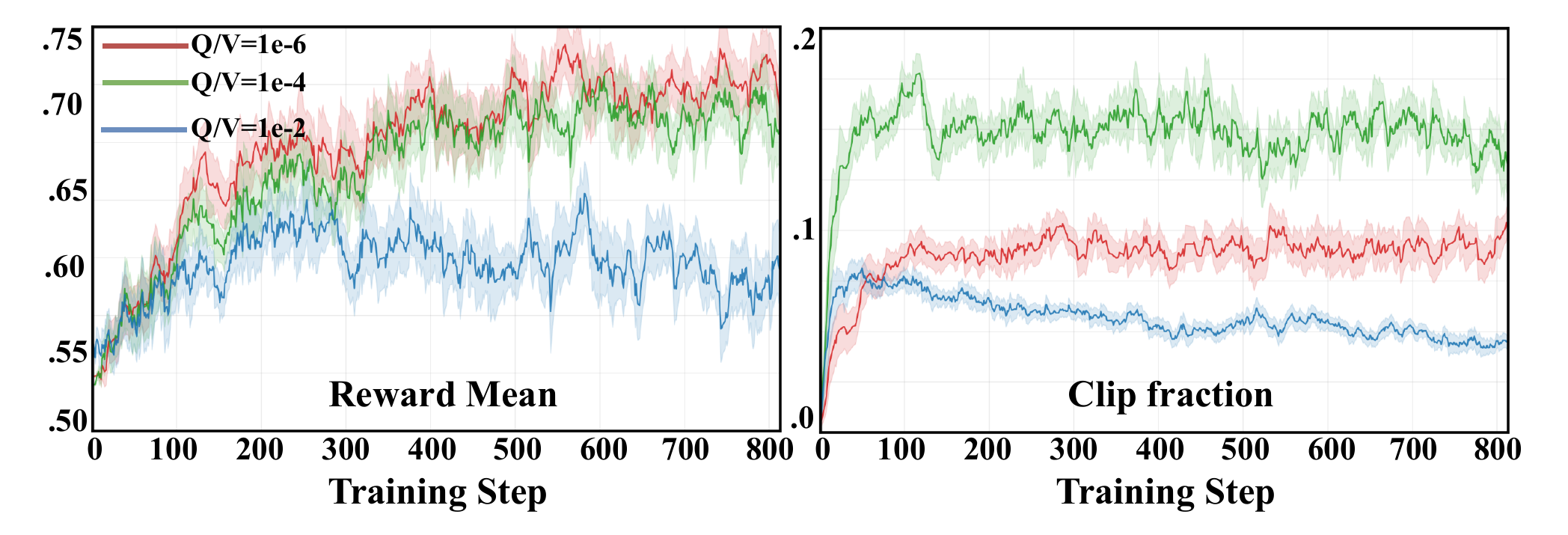}
       \vspace{-0.1in}
    \caption{Effect of the Kalman filter noise ratio $Q/V$ on training dynamics. 
    From left to right, we report the mean episodic reward and the PPO clip fraction. 
    Solid lines show the mean over multiple runs, and shaded regions indicate standard deviation.}
    \label{fig:parameter}
\end{figure}
\subsection{Parameter analysis}
In this section, we will analyze the effect of different values of $Q/V$ in \methodname{}, which controls the strength of temporal smoothing. We set three settings: $Q/V$ equals to $\mathrm{1e-6}$, $\mathrm{1e-4}$, and $\mathrm{1e-2}$. Empirically, a smaller $Q/V$ enforces stronger smoothing and higher temporal coherence, while a larger $Q/V$ makes the filter more responsive to short-term fluctuations. The experimental results, in terms of mean training reward and clip fractions, are shown in Figure~\ref{fig:parameter}.

\noindent\textbf{Reward mean.}\quad
From Figure~\ref{fig:parameter}, we can see that smaller $Q/V$ values consistently achieve higher rewards. The setting $Q/V=\mathrm{1e-6}$ yields the best final performance and shows stable improvement throughout training. $Q/V=\mathrm{1e-4}$ performs slightly worse but follows a similar trend. In contrast, $Q/V=\mathrm{1e-2}$ results in significantly lower rewards and exhibits performance degradation after mid-training, indicating that weak filtering introduces excessive noise into the ratio signal.

\noindent\textbf{Clip faction.}\quad
On the other hand, the clip fraction reflects the magnitude of policy updates. $Q/V=\mathrm{1e-6}$ maintains a moderate and stable clip fraction, suggesting balanced updates. $Q/V=\mathrm{1e-4}$ produces the highest clip fraction, indicating more aggressive updates that may risk instability. $Q/V=\mathrm{1e-2}$ leads to relatively low clip fractions, which is somewhat counterfactual. We conjecture that this may be because, in this case, the Kalman-filtered IS ratios may suffer from some structural estimation deviation.

Overall, the results highlight a clear trade-off between stability and effectiveness. With stronger filtering (smaller $Q/R$), the token-level ratios exhibit greater temporal coherence and training becomes more stable, leading to higher training rewards. By contrast, excessively large $Q/R$ weakens the filter, allowing high-frequency noise to leak back into the estimates, which degrades learning dynamics and final performance. In practice, $Q$ and $R$ should be tuned to reflect prior knowledge about ratio variability. For example, when using MoE models or when train--inference mismatch arises across rollout and inference engines, the underlying IS ratios may fluctuate more rapidly, and increasing $Q$ helps the filter track these faster variations. Conversely, when off-policy deviation is mild, a smaller $Q$ is preferable to impose slower, stronger smoothing.
  
\section{Conclusion}
We pioneer a new temporal view for token-wise importance-sampling ratios and empirically reveal a practical counterfactual mode in off-policy GRPO training: token-level importance-sampling ratios exhibit weak local coherence of off-policy deviation, which could amplify gradient variance, distort the policy-gradient updates of adjacent tokens and thus result in optimization instability. We propose \methodname{}, a causal Kalman filter over within-sequence ratios that treats ratio estimation as online state-space inference. \methodname{} suppresses noisy spikes while retaining the structural coherence of importance-sampling ratios, thereby improving both stability and effectiveness. Experiments on challenging math-reasoning benchmarks the superiority of \methodname{}, while remaining lightweight and compatible with existing pipelines. Since \methodname{} needs to process the sequential token autoregressively, it is difficult to parallelize the Kalman filtering procedure like the autoregressive generation of tokens. In the future, it is also interesting to design parallelized Kalman filtering algorithms.        


\bibliography{example_paper}
\bibliographystyle{icml2026}

\newpage
\appendix
\onecolumn

\section{Algorithm}
\subsection{The pseudo-code of \methodname{}}
\label{app:algorithm}

Algorithm~\ref{alg:kpo} summarizes the training procedure of \methodname{} at the trajectory level. 
The method first computes token/action-wise log importance ratios $z_t=\log \pi_\theta(a_t\mid s_t)-\log \pi_{\text{old}}(a_t\mid s_t)$, which serve as noisy observations of an underlying, smoothly varying log-ratio process. A one-dimensional Kalman filter is then applied in log space, using the process noise $Q$ and measurement noise $R$ to control the strength of temporal smoothing. Specifically, the prediction step propagates the posterior mean and variance of the latent log-ratio, while the update step corrects the prediction using the innovation $\delta_t$ and the Kalman gain $K_t$. The filtered estimates are finally mapped back to ratio space via exponentiation, yielding $\widetilde r_t=\exp(\hat{\rho}_{t\mid t})$, which is subsequently used to evaluate the \methodname{} objective in Eq.~(\ref{eq:J_kpo}). This design preserves the standard importance-ratio form while reducing high-frequency variability that otherwise amplifies gradient noise in policy optimization.

\begin{algorithm}[t!]
\caption{\methodname{}: Online Causal Kalman Filtering for Stable and Effective Policy Optimization}
\label{alg:kpo}
\begin{algorithmic}[1]
\REQUIRE Behavior policy $\pi_{\text{old}}$, updated policy $\pi_\theta$, initial posterior, process/measurement noises $(Q,R)$, trajectory $(s_{1:T},a_{1:T})$
\ENSURE The policy loss $\mathcal{J}_{\text{KPO}}$
\\
\STATE \textbf{Raw log-ratio computation:}

\STATE $z_t \leftarrow \log \pi_\theta(a_t\mid s_t)\;-\;\log \pi_{\text{old}}(a_t\mid s_t)$ \COMMENT{token/action-level log IS ratio}\\

\STATE \textbf{Kalman filtering in log space.} \COMMENT{1D local-level model: $\rho_t=\rho_{t-1}+\eta_t$, $z_t=\rho_t+\epsilon_t$}
\FOR{$t=1$ \textbf{to} $T$}
    \STATE \textit{Predict:}
    \STATE $\hat{\rho}_{t\mid t-1} \leftarrow \hat{\rho}_{t-1\mid t-1}$ 
    \STATE $P_{t\mid t-1} \leftarrow P_{t-1\mid t-1} + Q$ \COMMENT{Eq.~(\ref{eq:kalman_pred})}
    \STATE \textit{Innovation and gain:}
    \STATE $\delta_t \leftarrow z_t - \hat{\rho}_{t\mid t-1}$
    \STATE $S_t \leftarrow P_{t\mid t-1} + R$
    \STATE $K_t \leftarrow P_{t\mid t-1} / S_t$
    \STATE \textit{Update:}
    \STATE $\hat{\rho}_{t\mid t} \leftarrow \hat{\rho}_{t\mid t-1} + K_t \delta_t$
    \STATE $P_{t\mid t} \leftarrow (1-K_t)\,P_{t\mid t-1}$ \COMMENT{Eq.~(\ref{eq:update_block})}
\ENDFOR
\\
\STATE \textbf{Back to ratio space.}
\FOR{$t=1$ \textbf{to} $T$}
    \STATE $\widetilde{r}_t \leftarrow \exp\!\left(\hat{\rho}_{t\mid t}\right)$ \COMMENT{filtered IS ratio}
\ENDFOR
\\
\STATE \textbf{Objective.} Compute the policy loss $\mathcal{J}_{\text{KPO}}$ using $\{\widetilde{r}_t\}_{t=1}^T$ according to Eq.~(\ref{eq:J_kpo}).
\end{algorithmic}
\end{algorithm}

\section{The experimental setting}

\subsection{The experimental setting for Figure 1}
\label{app:off-policy}

To explore off-policy patterns, we run GRPO on Qwen3-4B and record token-level importance sampling (IS) ratios for 960 samples at the 100-th training step, following the original GRPO setup. We use a window size of 50 and a maximum sequence length of 4096 tokens. The training batch size is 32, and the mini-batch size is 8, resulting in four mini-batches per batch. Notably, only the first mini-batch is on-policy, and the remaining three mini-batches are off-policy (we only use off-policy samples). Other experimental setting is same to the main experiment setting shown in \ref{app:exp_set}. 

\paragraph{Window-wise off-policy frequency analysis.}
Using non-overlapping windows of length $50$ (up to $80$ windows per sample), we first restrict each sequence to valid positions with $\texttt{padding\_mask}=\texttt{True}$. The valid prefix of length $n_w\cdot 50$ is reshaped into $[n_w,50]$. For each window, we count $\texttt{up}=\sum \mathbb{I}[\texttt{log\_ratio}>0]$ and $\texttt{down}=\sum \mathbb{I}[\texttt{log\_ratio}<0]$, and define the off-policy frequency as $(\texttt{up}+\texttt{down})/50$. Finally, for each window index, we aggregate frequencies across all samples that contain that window using $\texttt{nanmean}$ and $\texttt{nanvar}$ to produce mean/variance curves for plotting.

\paragraph{Run-length computation.}
Run-lengths are computed per sample on valid tokens only ($\texttt{padding\_mask}=\texttt{True}$). Each valid position is first assigned a type $c_t\in\{\texttt{up},\texttt{down},\texttt{on}\}$ (by the sign of $\texttt{log\_ratio}_t$ or predefined ratio bins). For each type, the valid sequence is scanned left-to-right to record lengths of maximal contiguous runs: a counter $\ell$ is incremented while $c_t$ remains unchanged, and when the type differs the current $\ell$ (if $\ell>0$) is appended and then reset; the final run is appended if nonzero.

\paragraph{Switch-frequency computation.}
Switch frequency is computed in as follows: (i) restrict to valid tokens with $\texttt{padding\_mask}=\texttt{True}$ and map each position to a type $s_t\in\{-1,0,1\}$ (\texttt{down}, \texttt{on}, \texttt{up}); (ii) partition the valid type sequence into non-overlapping windows of length $50$; (iii) for each window $w$ of length $L$, count adjacent type changes $\sum_{j=1}^{L-1}\mathbb{I}[w_j\neq w_{j-1}]$ and normalize by $L-1$ to obtain a window-level switch rate; (iv) report the sample-level switch frequency as the mean over windows, falling back to the full valid sequence if fewer than $50$ tokens are available; (v) if the valid length is $\leq 1$, the switch frequency is set to $0$.

\subsection{The details of experiments}
\label{app:exp_set}

For the Math task, uniform hyperparameters are employed across all methods. The maximum response lengths are set to 4096 tokens, respectively. The actor learning rate is fixed at $1 \times 10^{-6}$. We employ group-based rollouts with a group size of 8. A binary rule-based reward function is used (1 for success, 0 for failure). The batch sizes for training and evaluation are 32 and 64, respectively. The loss aggregation mode is``sequence-mean-token-mean". During evaluation, we use nucleus sampling with $\text{top\_p}=1.0$ and a temperature of 1.0. For comparing methods, we use the suggested parameters in their papers. Specifically, the clipping parameter $\epsilon=0.2$ in GRPO and $\epsilon=0,4$ in GMPO. The clipping parameter $\epsilon^{-}=0.0003$ and $\epsilon^{+}=0.0004$ in GSPO. All experiments are conducted on NVIDIA H100 GPUs.

\subsection{The meanings of training metrics}
\label{app:train_metric}

\begin{itemize}
    \item \emph{Policy Gradient Loss:} Primary optimization signal. Smooth, gradual decrease suggests stable learning; sharp spikes or large magnitudes indicate overly aggressive updates and potential instability.

    \item \emph{Policy Gradient Clip Fraction:} Fraction of updates affected by clipping. Moderate levels suggest controlled steps; persistently high fractions imply frequent extreme updates and unstable optimization dynamics.

    \item \emph{Entropy:} Tracks policy stochasticity. Adequate entropy supports exploration and prevents premature collapse; rapid entropy decay indicates mode collapse or over-confident generation, while excessively high entropy can slow convergence.

    \item \emph{Mean Reward:} Average return per rollout. A smooth upward trend reflects effective learning; abrupt drops typically signal instability or reward hacking.
\end{itemize}

\section{Experimental analysis}

\subsection{Token proportion}
\label{app:token_pro}

Token-type assignment is based on the IS ratio. Before Kalman filtering, we treat tokens with an on-policy ratio $r=1$ as \emph{on-policy}, tokens with $r>1$ as \emph{up-policy}, and tokens with $r<1$ as \emph{down-policy}. After Kalman filtering, the IS ratios are smoothed, so tokens with exactly $r=1$ become rare. We therefore define a token as on-policy if $r\in[1-\epsilon^{-},\,1+\epsilon^{+}]$, with $\epsilon^{-}=0.0003$ and $\epsilon^{+}=0.0004$. Tokens with $r>1+\epsilon^{+}$ and $r<1-\epsilon^{-}$ are classified as up-policy and down-policy, respectively. As shown in Table~\ref{tab:raw_vs_filtered_metrics_grouped_transposed}, Kalman filtering reduces the measured fraction of on-policy tokens and increases the fraction of off-policy tokens. This shift should not be viewed as undesirable for policy optimization. It largely reflects the tighter on-policy band after smoothing, while the resulting off-policy ratios are no longer dominated by noisy spikes but become structurally coherent. Consistently, the token run-length and switching-frequency statistics indicate that local off-policy deviations remain smooth and stable, suggesting that the induced token-type distribution is well-behaved.

\subsection{The frequency-domain analysis}
\label{app:frequency}

\noindent\textbf{Low-frequency ratio.}\quad
Beyond time-domain characterizations of temporal structure, we study the IS ratio sequence in the frequency domain. Let $\{r_t\}_{t=0}^{T-1}$ denote the token-level ratio sequence (in our implementation, $r_t$ may refer to the log-ratio), and let
$\bar r \;=\; \frac{1}{T}\sum_{t=0}^{T-1} r_t,\qquad x_t \;=\; r_t-\bar r $ be the mean-centered signal, so that the DC component does not dominate the energy accounting. Define the discrete Fourier transform (DFT) $X_k \;=\; \sum_{t=0}^{T-1} x_t \, e^{-j\frac{2\pi}{T}kt},\qquad k=0,1,\dots,T-1,$ and the associated spectral power (periodogram) $P_k \;=\; |X_k|^2.$ Given a cutoff index $k_c \in \{0,\dots,\lfloor T/2\rfloor\}$, we define the low-frequency band as the union of symmetric bins
\[
\mathcal{K}_{\mathrm{LF}} \;=\; \{0,1,\dots,k_c\}\;\cup\;\{T-k_c,\dots,T-1\},
\]
which ensures that the low-frequency energy is counted consistently for real-valued sequences. We then define the \emph{low-frequency ratio} (LFR) by
\[
\mathrm{LFR}(k_c) \;=\; \frac{\sum\limits_{k\in \mathcal{K}_{\mathrm{LF}}} P_k}{\sum\limits_{k=0}^{T-1} P_k}.
\]
By construction, $\mathrm{LFR}(k_c)\in[0,1]$ quantifies the fraction of total signal energy residing below the cutoff frequency $\omega_c = 2\pi k_c/T$: small values indicate energy concentrated in high-frequency components (rapid token-to-token fluctuations), whereas large values indicate predominantly low-frequency variation over longer token spans. In our measurements, the unfiltered ratio sequence yields a low LFR of $0.12$, consistent with a spectrum dominated by high-frequency energy. After applying Kalman filtering, the LFR increases to $0.98$, implying that the filtered sequence is almost entirely explained by low-frequency components. This behavior is consistent with the role of the Kalman filter as a temporal denoiser: it suppresses fast oscillations while preserving slowly varying structure.


\end{document}